# Make me an Offer: Forward and Reverse Auctioning Problems in the Tourism Industry

Ioannis T. Christou[1,2], Dimitris Doukas[3], Konstantina Skouri[4], Gerasimos Meletiou[4]

[1] The American College of Greece, Athens, Greece

[2] Net Company-Intrasoft, Luxembourg

[3] Twinnet Information Systems Sole Proprietor Ltd, Athens, Greece

[4] University of Ioannina, Ioannina, Greece

1. Abstract

Most tourist destinations are facing regular and consistent seasonality with significant economic and social impacts. This phenomenon is more pronounced in the post-covid era, where demand for travel has increased but unevenly among different geographic areas. To counter these problems that both customers and hoteliers are facing, we have developed two auctioning systems that allow hoteliers of lower popularity tier areas or during low season periods to auction their rooms in what we call a forward auction model, and also allows customers to initiate a bidding process whereby hoteliers in an area may make offers to the customer for their rooms, in what constitutes a reverse auction model initiated by the customer, similar to the bidding concept of priceline.com. We develop mathematical programming models that define explicitly both types of auctions, and show that in each type, there are significant benefits to be gained both on the side of the hotelier as well as on the side of the customer. We discuss algorithmic techniques for the approximate solution of these optimization problems, and present results using exact optimization solvers to solve them to guaranteed optimality. These techniques could be beneficial to both customer and hotelier reducing seasonality during middle and low season and providing the customer with attractive offers.

2. Introduction

Prior to the pandemic, tourism was one of the world's largest sectors, accounting for 1 in 4 of all new jobs created in the world, 10.3% of all jobs (333 million), and 10.3% of global GDP (World Travel & Tourism Council (WTTC) annual Economic Impact Research). Travel & Tourism enables socio-economic development, job creation and poverty reduction. The effect of COVID-19 emphasized the importance and contribution of tourism. In 2020, 62 million jobs were lost, leaving 271 million employed across the sector globally. While 2021 saw the beginning of the recovery for the global tourism sector, this was slower than expected. However, according to WTTC, the future outlook is positive and tourism GDP could return to 2019 levels by the end of 2023 expecting to create nearly 126 million new jobs within the next decade outpacing the growth of the overall economy (2.7% per year).

Seasonality in tourism is a well-known phenomenon that affects every tourist destination, prompting hotelier to effectively plan the use of their resource. According to Baron (1975) there are two distinct types of seasonality "natural" and "institutional". The four regular seasons (Spring, Summer, Autumn, and Winter) influence are known for being "natural" causes of seasonality. While, "institutional" seasonality is also a well-known cause of seasonality created by human social, political, and economic agencies. So, the Holy Days, such as Christmas, Easter, Ramadan, etc. as well as summer school holiday period is believed to be the leading institutional cause of tourist seasonality. The institutional seasonality has a long history in time which makes it more challenging to change these well-established patters. Figure 1 represents the high seasonality effect in Greece during 2022. Greece is a country that according to data released annually by the Association of Greek Tourism Enterprises (SETE) and the Bank of Greece, tourism is a major contributor to Greek economy accounts for 18% of Greece's GDP and employs more than 900,000 people, accounting for one fifth of the workforce.

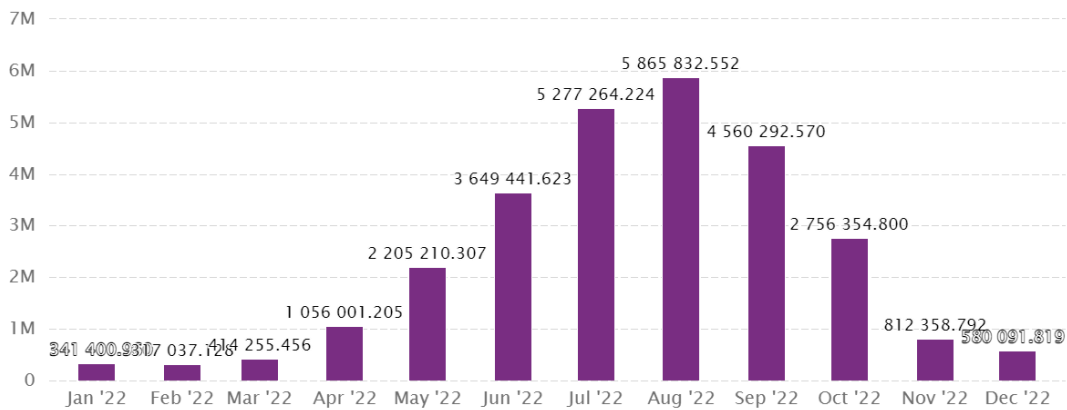

Figure 1. International tourist arrivals in Greece during 2022 (source https://www.unwto.org/tourism-data/global-and-regional-tourism-performance)

### 2.1. Related Work

Tourism seasonality has social, economic and environmental effects, including price increases during the high season, income and employment instability with repercussions on commitment and the quality of the services offered (Fernández-Morales and McCabe 2016; Gyða and Rögnvaldur 2017; Rosselló and Andreu 2017; Suštar and Laškarin Ažić 2020; Bampatsou et al. 2022). Furthermore, the degree of seasonality constitutes one of the major factors with significant impact on hotel efficiency (Lado-Sestayo and Fernández-Castro, 2019). However, a main seasonality characteristic is that it is regular and consistent, thus it can be predictable and so more possible to be mitigated (Christou, 2011).

Eliminating seasonality can be a difficult task, since tourism is characterized by freedom of choice, tourists likely not intend to alter their behaviors without attractive options. One way of mitigating the seasonality consequences is to find strategies for seasonal adjustment. This it can also extend the tourism season whilst helping to synergize ideas and increase business sales. In addition, the surge in "workation holidays" enabled by remote and flexible working offers new opportunities to travel providers and destinations. Furthermore, this has created a need for novel, innovative, more flexible booking systems that will allow hoteliers and the tourism industry to better deal with this new very high level of demand without increasing (in the short-term at least) its current capacities by too much. Although there are many challenges that tourist industry faces, this study could be applicable to mitigate seasonality.

In Wu et al. (2021), the authors studied several strategies for reducing churn and increasing repeated business from the same customers in the online room rental market. They relied on social interaction theory and a longitudinal dataset to draw conclusions about the major attributed that drive customers to repeatedly do business with the same room provider throughout a number of years. Even though price turns out to be a significant attribute for repeat business by customers, auctions are not mentioned in that study, as they are more likely to create new revenue streams instead.

On the other hand, Ding, Gao and Liu (2022) showed that positive online reviews have a strong positive effect on the growth of new independent (therefore relatively small) hotels, and vice-versa, negative reviews have a strong negative effect on the growth of such hotels.

In this context, we propose to exploit the potential of mobile and optimization technologies in a double-auction scheme simultaneously implemented in a fully developed Decision Support System (Turban, Aronson, Liang, 2005) as follows: (1) forward auctions initiated by hoteliers for their resources (rooms), for which potential travelers may bid on, (2) reverse auctions initiated by potential travelers interested in visiting an area, for which the hoteliers of the relevant region may place offers for their rooms (reverse bids). We show that each type of auction has unique advantages (as well as disadvantages) and we show how hoteliers and/or travelers can benefit from each auction type.

Forward auctions, initiated by hoteliers, is studied in Marentakis & Emiris (2010); their focus however is on designing an architecture for providing mobile auctions using location-based services. There is no discussion on the underlying optimization problems that the hoteliers will face in order to maximize their profits given a set of bids from potential customers.

Bichler and Kalagnanam (2005) study the problem of how to optimally configure an offer made as a response to an essentially reverse auction in our setting, and in addition model as a Mixed Integer Programming problem, the problem of choosing the reverse auctions to respond to, given the supplier's limited resources and competition from other suppliers. While in our work we also explicitly assume multiple hoteliers will respond to any given reverse auction, we only specify the methodology by which the hotelier can decide what rooms and amenities to include in their offer, and also provide a specialized data mining algorithm for the estimation of the expected winning price bid, from the hotelier's perspective.

Both types of auctions we propose differ significantly from any existing room reservations systems such as Trivago, Booking.com, or match-making sites such as AirBnB. Our reverse auctioning system resembles the defunct as of 2020 bidding feature of priceline.com where potential guests could initiate a bid for a hotel room in a particular geographic area (enclosed within a pre-specified perimeter not editable by the user), however it differs significantly in both how the visitor can specify their preferences, and even more importantly in how the hotel prices are specified. In priceline.com, the rooms to be auctioned would be drawn from a pre-specified pool of rooms whose price would also be determined at a prior time; the difference in price between a winning bid and the (low predetermined) price asked by the hotel would determine the revenue of priceline.com. In this practice (known as *opaque selling*, see Chen, Gal-Or and Roma (2014), and more recently Ren and Huang (2021)), the visitor is not aware of the exact room they are renting until the transaction has been fully implemented and paid for; the rooms to be rented belong to a special "left-over" category as far as the hotelier is concerned.

In our system, the hoteliers are free to monitor the initiated reverse auction of every user and respond dynamically to each such auction by differentiating both prices and amenities specified for the offered room. We present data mining-based tools to help the hotelier make reasonable estimates of the expected price they should ask in their offering to have strong chances of winning the auction. Further, whereas the bidding feature of priceline.com was a ubiquitous feature available in any area, we specifically target less popular areas where both supply and demand are typically much sparser than the more popular tourist areas.

Our proposed auctions offer a new model for optimal match making between potential travelers and resources (hotel rooms etc.) through the advanced modeling and algorithmic management of the bids that are made through the system. Our models allow the maximization of hotelier profits through multiple paths (income maximization from existing bids, minimization of vacant room-days where at least one available room remains empty due to bad scheduling of customers' arrivals/departures and so on), and at the same time allow for the simultaneous customer service level maximization as depicted by the total number of offers satisfied etc.

## 3. Classical Auction and Its Variations in Multi-Object Auctioning

Classical auctioning (Roth and Sotomayor, 1990) of an object works as follows: a seller decides to auction an object of some monetary value that is generally unknown and in general different from person to person. On a specified date, the auction for the object begins, with the seller specifying an initial price for the object. Following this, potential buyers are allowed to express their willingness to purchase the object in this price. If there are no buyers in the first round, the auction closes (fails) and no transaction takes place. If there is just a single potential buyer at the stated price, the object is allocated to them for the stated price, and the auction closes with a successful outcome. For as long as there are at least 2 interested buyers, the seller keeps increasing the price (usually by a fixed price increment amount), and keeps doing so until there is only a single candidate left willing to purchase at the current price, which becomes the price the buyer *must* pay the seller. Needless to say, there are many variations on the above auctioning scheme which is better known as "English auction".

In this work, we consider forward auctioning schemes where hoteliers may auction any number of their hotel rooms they want, by specifying a booking period, the start and end dates of the auction (the "auction active period"), and an initial minimum daily rental price for each room type, below which the room cannot be rented. Potential customers may specify once their unique bid for any of the rooms being auctioned for any sub-period of the booking period specified and will know if they have won the auction only after the end of the active auction period. As we shall see shortly, because of the characteristics of the forward auction scheme we consider, there is no guarantee for the buyers that the highest price offered will be among the winners of the auction, as the selection of the winners depends on other characteristics of the bid such as the dates for which they place their bid. As an example, a hotelier could specify an auction for 5 double-bed rooms in their hotel, with sea-view, American breakfast included, for the period June 1 — 15 2023, with an initial starting price set to €65/night, with the auction starting on Jan. 1st, 2023 and ending on May 1st, 2023. Potential customers can specify their bid during this period for any or all of the rooms auctioned during this period. An interested customer may specify only once, at any time between Jan. 1st and May 1st, that they are interested in making a reservation for 1 of the auctioned rooms in the hotel for any consecutive three nights during the specified period June 1 — 15, 2023 at the minimum suggested price of €65/night. Another customer, less flexible in their arrival and departure time but willing to bid higher per night, could specify a bid that asks for two nights in the period June 10 —12 2023, at the price of €75/night.

Once the active auction period is over, the hotelier has received all bids from interested customers. One (sub-optimal) way to choose bids to accept would be to sort all bids in descending order of offered price per night, and then use as secondary sort key, descending order of total requested nights of stay; then keep accepting bids from the sorted list as long as there is no conflict with already accepted bids until the end of the list. This greedy algorithm is in general better than sorting bids in ascending order of the time they arrive and then accepting from such a sorted list as long as there is no conflict with previously accepted bids which implements the standard First-Come, First-Served (FCFS) protocol that hoteliers traditionally implement. The reason is that it allows more profitable bids to be considered before less profitable ones are; still, it carries no guarantees as to the maximization of profits for the hotelier, as high-value bids will likely overlap in time with both other high-value as well as lower-value bids, and choosing the best combination of bids to accept while avoiding any conflicts is far from trivial. We model this hotelier-induced forward auction problem then, as a Mixed Integer Programming problem.

### 3.1 Optimal Bid Selection in Forward Auctions
*Notation*

| | | |
|---|---|---|
| | $r$ | The type of room (e.g. 1-bed) for which the offer is made $r \in R_r = \{1,2,\dots,R\}$ |
| | $\pi_r$ | the minimum price for each room that the hotelier is willing to consider for a given bid, $r \in R_r$ |
| | $c$ | the customer who submitted the bid, $c \in C_r = \{1,2,\dots,C\}$ |
| | $n_r$ | the number of rooms of type $r$ that the hotelier has auctioned |
| | $D_c$ | set of dates during which the client $c$ wishes to visit hotel, so that $D_c \in [L_c, U_c]$ |
| | $M_c$ | denotes the total number of nights the client is bidding for |
| | $d_i$ | Date $i$ |
| | $x_{c,r,d} = \begin{cases} 1 & \text{if room} - \text{type } r \text{ is allocated to customer } c \text{ for the date } d \\ 0 & \text{else} \end{cases}$ | |
| | $y_c = \begin{cases} 1 & \text{if the bid by customer } c \text{ is accepted} \\ 0 & \text{else} \end{cases}$ | |
| | $l_c$ | integer variables denote the arrival date of customer $c$ if their bid is accepted, otherwise the value is arbitrarily set to any value in $[L_c, U_c]$. |
| | $u_c$ | The integer variables denote the departure date of customer $c$ if their bid is accepted, otherwise the value is set to any value in $[L_c, U_c]$. |

*Assumptions*

1) Let $R_r$ be the set of all room-types the hotelier auctioned in the current forward auction and let $n_r$ the number of rooms of type $r \in R_r$ that the hotelier has auctioned.
2) Let $\pi_r > 0, r \in R_r$ be the minimum price for each room that the hotelier is willing to consider for a given bid.
3) The period for which the hotelier auctions rooms in a given auction is continuous, and this entire period is a contiguous sequence $D = [d_1, d_2, \dots d_{|D|}]$.
4) Potential customers may submit their bids as *one or more tuples* of the form:
   $o = < c, r, n_{c,r}, D_c = \{d_{l_c}, d_{l_c+1}, \dots, d_{u_c}\} \subseteq \{L_c, U_c\} \subseteq D, |D_c| = M_c, b_{c,r,D_c} >$
   where $c$ denotes the customer who submitted the bid, $r$ is the room type (e.g. 1-bed) for which the offer is made, $n_{c,r}$ denotes the number of rooms of type $r$ the client wishes to book, $D_c$ denotes a variable set of dates during which the client wishes to visit that are bound from below by the earliest possible date $L_c$ and from above by the latest possible departure date $U_c$. The parameter $M_c$ denotes the total number of nights the client is bidding for, and the quantity $b_{c,r,D_c} \geq \pi_r$ denotes the bid value (per night) of the given customer for the given room type and period.
5) We consider that $\forall d \in D_c: b_{c,r,d} = b_{c,r,D_c}$ and that for every other date $d$, $b_{c,r,d} = 0$; the customer may submit more than one such tuples (offers) however they must all be for the same time-period, and each tuple must specify a different value for the room-type $r$. For any given customer, either all of their offers comprising the bid must be accepted by the hotelier, or else none must be accepted.
6) There once the clients submit their initial bids, there is no further action required from them; they are either notified by the hotelier that their bid is accepted, in which case they are legally obliged to abide by their bid, or they are notified by the hotelier that their bid is rejected. In either case, the customers cannot engage in any further negotiations with the hotelier regarding this auction.

Based on the above assumptions, from the perspective of the hotelier who has initiated a semi-classical forward auction and received a number of bids for the rooms they auctioned, the objective is rather clear; the hotelier wishes to maximize their total profits by selecting the right combination of submitted bids.

Before specifying the full model, we need to define the index-function $I(d): D \to \mathbb{N}$ that simply returns the position of the date $d$ in which it appears in the sequence $D$ so that $I(d_k) = k$.

In this context, we formulate a first model (IP1) for the problem of the optimal bid selection by the hotelier, as follows.

$$(IP1) \max_{x,l,u,y} \sum_{c \in C_r, r \in R_r, d \in D_c} n_{c,r} b_{c,r,d} x_{c,r,d} \quad (1)$$

$$s.t. \begin{cases} \sum_{c \in C_r} n_{c,r} x_{c,r,d} \leq n_r \ \forall r \in R_r, \forall d \in D_c & (2) \\ \sum_{d \in D} x_{c,r,d} = y_c M_c \ \forall c \in C_r, \forall r \in R_r & (3) \\ (I(d) - |D_c|) x_{c,r,d} \geq l_c - |D_c| \ \forall c \in C_r, \forall r \in R_r, \forall d \in D_c & (4) \\ I(d) x_{c,r,d} \leq u_c \ \forall c \in C_r, \forall r \in R_r, \forall d \in D_c & (5) \\ L_c \leq l_c \leq u_c \leq U_c \ \forall c \in C_r & (6) \\ u_c - l_c \leq M_c - 1 + |D_c|(1 - y_c) \ \forall c \in C_r & (7) \\ l_c, u_c \in \mathbb{N} \ \forall c \in C_r & (8) \\ x_{c,r,d}, y_c \in \{0,1\} \ \forall c \in C_r, \forall r \in R_r, \forall d \in D_c & (9) \end{cases}$$

The objective function (1) of problem (IP1) simply calculates the hotelier total income from the acceptance of bids; the constraints simply make sure that this maximization occurs without any violation of resources (the customers whose bids are satisfied must be able to come to the hotel and find all the rooms they bid for available for them, and them only).

In particular, the constraints (2) specify that for each room-type $r$, the total number of rooms allocated to customers (by accepting their bids) is less than or equal to the total number of rooms for that type that were auctioned, and this is satisfied for every day of the auction period. This is essentially a capacity constraint.

The next set of constraints (3) specifies the relation between the variables $x, y$: if the customer $c$'s bid is rejected, then $y_c = 0$ and this forces $x_{c,r,d} = 0$ for all $r \in R_r, d \in D_c$, given the binary nature of the variables. On the other hand, if the customer's bid is accepted, then this means that for each room-type $r$ that was bidded, exactly $M_c$ of the variables $x_{c,r,d}$ will take the value 1. The exact $x$ variables that will take on the value 1 are specified with the help of the constraints (4)—(7).

The constraints (4) ensure that the variables $x_{c,r,d}$ take on the value 1 only for those dates $d$ for which $l_c \leq I(d)$ (notice that when $x_{c,r,d} = 0$, the constraint becomes $l_c \leq |D_c|$ and is inactive since constraints (6) are more strict).

Similarly, the constraints (5) constrain the variables $x_{c,r,d}$ to take on the value 1 only for those dates $d$ for which $u_c \geq I(d)$ (notice that when $x_{c,r,d} = 0$, the constraint becomes $u_c \geq 0$ and is redundant).

Constraints (6) simply constraint the arrival and departure date of a customer to be within the period the customer's bid specified, and constraints (7) dictate (when considered together with constraints (3), (4) and (5)) that if a customer's bid is accepted, then the departure and arrival date of the customer will be such so that the customer stays for exactly the number of nights they specified in their bid ($M_c$). Constraints (8) and (9) simply specify the discrete nature of the decision variables.

### 3.2 Improving the Model (IP1)

The model (IP1) can be further simplified and improved so as to be more easily solved by modern optimization algorithms. As a first improvement, notice that we can drop the index $r$ from the variables $x_{c,r,d}$ to end up with a smaller set of $|C_r||D_c|$ binary variables $x_{c,d}$ $c \in C_r, d \in D_c$; this is because as long as a customer's bid is accepted, all their different room-type requests are guaranteed to be accepted, and if on the other hand their bid was rejected, then none of their different room-type requests was accepted.

In addition, the variables $u_c$ upper-bounding the dates for which the $x$ variables can be set to 1, are connected with the lower-bounding variables $l_c$ via the equation

$$u_c = l_c + M_c - 1 \ \forall c \in C_r.$$

Given the above, we can formulate the following equivalent model (IP1') that has significantly fewer discrete variables:

$$(IP1') \max_{x,l,y} \sum_{c \in C_r, r \in R_r} n_{c,r} b_{c,r} M_c y_c \tag{1}$$

$$s.t. \begin{cases} \sum_{c \in C_r} n_{c,r} x_{c,d} \leq n_r \ \forall r \in R, \forall d \in D_c & (2) \\ \sum_{d \in D_c} x_{c,d} = y_c M_c \ \forall c \in C_r & (3) \\ (I(d) - |D_c|) x_{c,d} \geq l_c - |D| \ \forall c \in C_r, \forall d \in D_c & (4) \\ I(d) x_{c,d} \leq l_c + M_c - 1 \ \forall c \in C_r, \forall d \in D_c & (5) \\ L_c \leq l_c \leq U_c + 1 - M_c \ \forall c \in C_r & (6) \\ l_c \in \mathbb{N} \ \forall c \in C_r & (7) \\ x_{c,d}, y_c \in \{0,1\} \ \forall c \in C_r, \forall d \in D_c & (8) \end{cases}$$

The above model (IP1´) can be further improved with regards to the following:

1. Potential customers can easily express their interest to visit the hotel during certain non-contiguous date intervals within the auction period; this would be mostly of interest to so-called weekend tourists: if a customer is interested in visiting during the first or second weekend of a period, this can be easily modeled by describing a bid by the customer for all room-types they wish to book for the period covering all days they are interested in staying, providing the exact total number of nights they want to stay, and adding the constraints $x_{c,d} = 0$ for every day $d$ during which they are not interested in staying.
2. Regarding the objective function, that currently computes income maximization for the hotelier, we can easily modify it to express total profits maximization; as long as we know for every room-type $r$ its daily operating cost $\kappa_r \geq 0$, then the profit maximization objective can be written as

$$\sum_{c \in C_r, r \in R_r} n_{c,r}(b_{c,r} - \kappa_r) M_c y_c$$

and the only difference from the original model (IP1) is in the coefficients of the variables $y_c$ in the objective function.

3. Finally, consider the case where different room-types in reality express the same room with different amenities, so that for example, room type $r_1$ corresponds to a sea-view 2-bed room with American breakfast included, whereas room type $r_2$ corresponds to the same room but without any breakfast included. We can still model the problem by "upgrading" the status of the parameters $n_r, r \in R_r$ to integer model *decision variables* indexed by date to become $n_{r,d}, r \in R_r, d \in D_c$ —not to be confused with the parameters $n_{c,r}$ that the customer specifies in their bid— because of the possibility of every such "virtual" room-type to change according to which customers are staying in the hotel every day. The following extra coupling constraints ensure that real hotel room capacity is not exceeded under any circumstances:

$\sum_{r \in R_q} n_{r,d} \leq N_q \quad \forall d \in D_c, \forall q \in Q$  (9)

where the set $Q$ contains indexes for every "real" room-type, the values $N_q$ express the total number of rooms of every "real" room-type, and the sets $R_q$ contain the indexes corresponding to every real room-type. For example, if the hotel has 5 one-bed rooms and 10 two-bed rooms, each of which can be booked with American, Continental, or no breakfast, then the hotelier has 6 room-types in total: 3 "virtual" room-types for the one-bed rooms, and another 3 "virtual" room-types for the two-bed rooms. In this case then, $Q = \{1,2\}, N_1 = 5, N_2 = 10$, and the sets $R_q$ are as follows: $R_1 = \{1,2,3\}, R_2 = \{4,5,6\}$. The variables $n_{1,d}, n_{2,d}, n_{3,d}$ correspond to the number of one-bed rooms that are booked on day $d$ with American ($n_{1,d}$), Continental ($n_{2,d}$) or no-breakfast ($n_{3,d}$); finally, the variables $n_{4,d}, n_{5,d}, n_{6,d}$ correspond to the number of two-bed rooms that are booked with American ($n_{4,d}$), Continental ($n_{5,d}$), and no-breakfast at all ($n_{6,d}$) respectively.

## 4. Reverse Auctions for Hotel Rooms

Reverse auction is a conceptually simple variant of the classical forward auction scheme, that has become commonplace practice especially among public sector and private companies. It has since proven equally effective for Business-to-Business (B2B) transactions, and we propose that it can have significant value in Consumer-to-Business (C2B) transactions as well. In a reverse auction the consumer expresses their interest in a specific product or service, and potential business providers submit their (reverse) bids in response to the consumer request. The auction is open to submissions for a given period of time, after which no bids are accepted any more. At this point, the provider who submitted the lowest price bid wins the reverse auction, though the price the winner pays may sometimes be not the price the provider asked for, but the second-lowest price submitted instead (this helps a lot public projects be carried out with minimal risk regarding budget being too low for the project.)

In situations where demand is low, reverse auction schemes may lead to better prices for the consumer as the service providers compete against each other over winning the customer's business. In reverse auctions the customer is seldom faced with any complex decision making problem leading to an optimization model formulation; the customer will usually end-up picking the lowest priced bid among the ones that satisfy their needs; more complex sets of criteria can easily be incorporated in the customer's decision making process as simple sort-keys in a *sorting algorithm* that eventually produces an ordering of the bids for the customer to select the top offer.

## 4.1. Bid Optimization in Reverse Auctions

From the point of view of the hotelier that is considering submitting a bid in response to a new reverse auction initialized by a potential customer, the decision they have to make is about the price they are going to give in their offer so that it maximizes hotelier's expected profits. Clearly, the higher the price the hotelier asks, the more their final profits; on the other hand, the higher the price the hotelier asks, the smaller the probability that the customer will choose the hotelier's offer among all possible offers in that reverse auction. The expected profits for the hotelier can be computed as long as the hotelier can accurately estimate the customer's so-called "reserve price" which is the maximum price the customer may be willing to pay for the service of the customer. But even if the customer explicitly provides this information up-front (as a maximum price that would be accepted by them) before the reverse auction has begun, the hotelier has to estimate their competitor's offers. Here is why: in such reverse auctions, the hotelier's optimal strategy (Salant, 2014) is to set their bid price offering at exactly their real cost (including desired profit margins). By setting any lower price, the hotelier will end up losing money if their offer is accepted, while higher prices will tend to lose the auctions in the long run (unless all the competition has higher cost structure than the given hotelier).

Despite the above, the hotelier stands to gain significant benefits by an accurate estimation of the customer's "reservation price" for this auction, because, knowing the maximum price the customer is willing to pay gives the hotelier significant leverage in determining their auction bid price offer, especially when the customer's reservation price is higher than the hotelier's cost. But even in the case where the customer's reservation price is lower than the hotelier's cost, this knowledge is important so that the hotelier can save their time and efforts related to the submission of bids in such losing for them reverse auctions.

### 4.1.1. Customer Reservation Price Estimation in Reverse Auctions

It is obvious then, that accurate estimations of specific customers' reservation price is of great importance to the hoteliers being able to make optimal decisions for their participation in such reverse auctions. For example, if the hotelier knows that a customer's reservation price $\rho > 0$ is higher than their cost $\kappa > 0$ they can choose any price in the interval $[\kappa, \rho]$ and be certain that if their offer is rejected, it will only be because of a better offer from a competitor (this is why in the long-term, the hotelier's best strategy is to set their bid at exactly $\kappa$.)

Next, we model the problem of the customer's reservation price estimation as a Machine Learning problem; in particular, we consider the reservation price of a customer in every auction to be a random variable drawn from a probability distribution that we try to estimate as accurately as possible given data from previous reverse auctions.

Assuming that there is a dataset containing all reverse auction offers $\pi_1 \leq \pi_2 ... \leq \pi_N$ that were accepted in other reverse auctions that specified the same criteria as the current one, we can define the distribution of the accepted price (not exactly the same as the reservation price of the customer) as the following discrete distribution:

$$F(X \leq \pi_i) = \frac{argmax\{j:\pi_j=\pi_i\}}{N}, i = 1,2 \ldots N \quad (10)$$

| Auction # | Auction Bids | Accepted Offer Price |
|---:|---|---:|
| 1 | 30,50,100 | 30 |
| 2 | 40,50,60 | 40 |
| 3 | 40,40,50 | 40 |
| 4 | 40,50,80 | 40 |
| 5 | 45,50,55 | 45 |
| 6 | 45,60,60 | 45 |
| 7 | 48,55 | 48 |
| 8 | 50,60,60,80 | 50 |
| 9 | 45,50,70,80 | 50 |
| 10 | 50,60 | 50 |

*Table 1: Example of Reverse Auction Offers*

Consider the following toy example of a dataset of previous successful reverse auctions shown in Table 1. From this data, it is easy to produce the probability mass function of the random variable $X$ representing the accepted price of the reverse auction, illustrated in fig. 1 below.

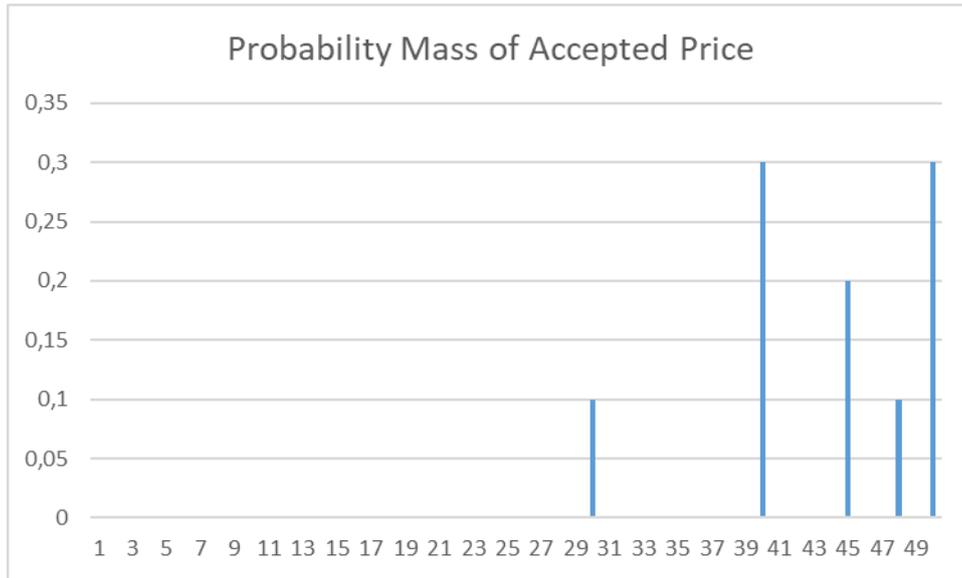

*Figure 1: Probability Mass of Accepted Price of the Example Dataset in Table 1*

And from fig. 1, the probability distribution function is immediately derived and shown in fig. 2.

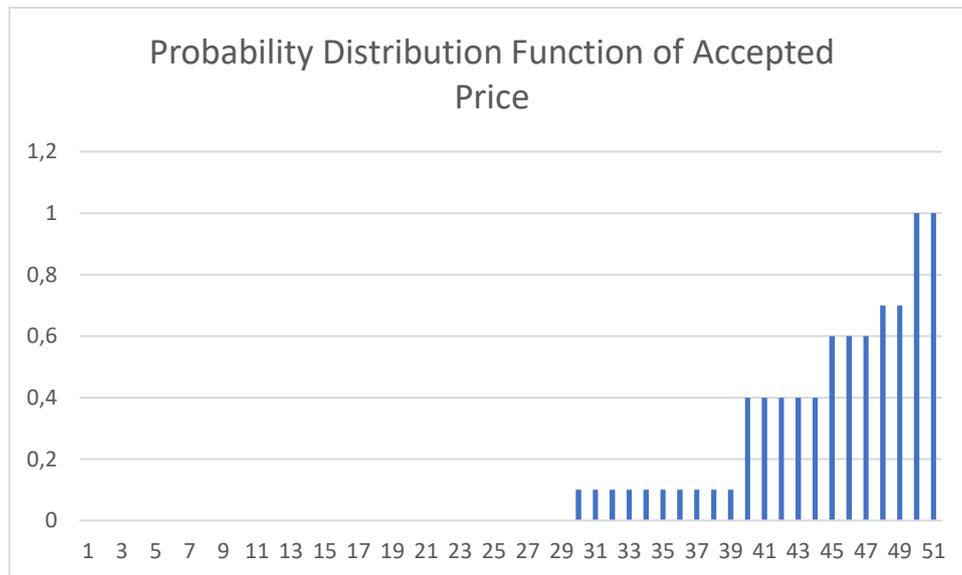

*Figure 2: Distribution Function of Accepted Price of the Example in Table 1*

Given the probability distribution of the acceptance price of such a reverse auction, the hotelier may easily derive the probability of acceptance for any offer they make for the current reverse auction. For example, by looking at the figure, the hotelier knows that an offer of less than or equal to €40 has at most 10% chance of rejection due to price.

When the estimation of the distribution of the acceptance price for a reverse auction is reasonably accurate, the hotelier can formulate and easily solve the problem of maximizing their expected profits in such a reverse auction as follows.

### 4.1.2. Hotelier Expected Profit Maximization in Reverse Auctions

Let's assume that the (variable) cost of the offer the hotelier is considering as a response on a reverse auction has been accurately computed to be $c > 0$. Given the probability mass $p_1, p_2, \ldots p_k$ satisfying:

$$\sum_{i=1}^{k} p_i = 1, \ p_i \geq 0, i = 1, \ldots k \tag{11}$$

Consider all the different accepted prices $\pi_1, \pi_2, \ldots \pi_k$ of a reverse auction that has occurred multiple times; the hotelier's expected profit $P$ for any given price $\pi > 0$ is then computed as follows:

$$E[P] = (\pi - c) \sum_{i:\pi_i \geq \pi} p_i \tag{12}$$

And to maximize their profits, the hotelier has to solve the following continuous maximization problem (MP):

$$\max_\pi (\pi - c) \sum_{i:\pi_i \geq \pi} p_i \tag{MP}$$

Solving (MP) can be done in time $O(k)$ where $k$ is the number of different price break points in the mass function.

*For the example in Table* 1, for a hotelier cost equal to $c = €10$ the price offer that maximizes the expected hotelier profit is at $\pi = €40$ at which point the expected profit reaches the maximum value of €27, as the reader can easily verify in fig. 3 from which we can easily deduce that an offer at €40 has a 90% chance of being accepted, so the expected profit becomes $(40 - 10) \times \frac{9}{10} + 0 \times \frac{1}{10} = 27$ where 0 is the profit of the hotelier if their offer is rejected.)

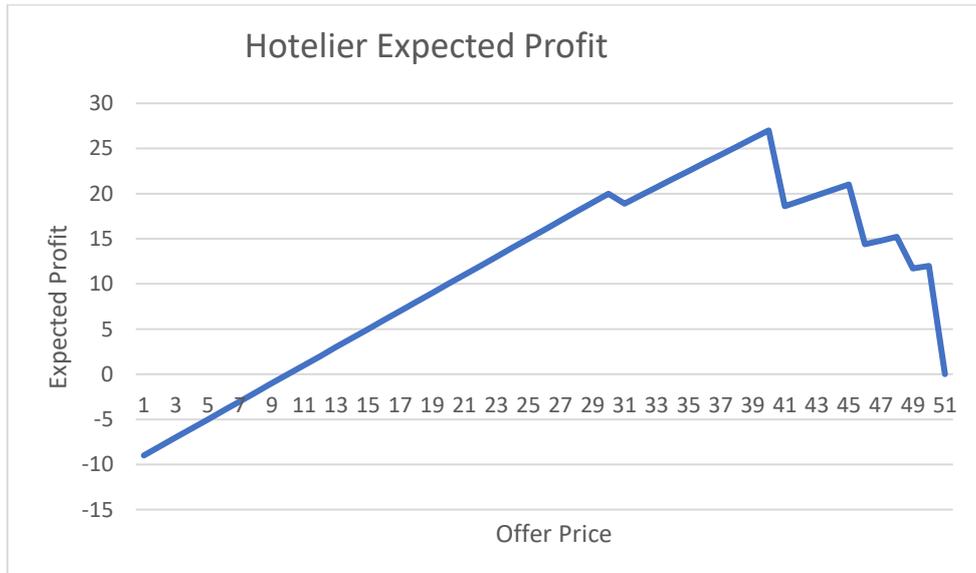

*Figure 3: Hotelier Expected Profit for the Example in Table 1 as a function of the offer price*

The above model seems to suggest that the Hotelier's problem of figuring out what the optimal price point is in response to a given reverse auction is a rather easy process. However, even if we do manage to get enough data for reverse auctions that have essentially the same characteristics, there is always the issue of the interpretation of the probability we are computing: in reality, the probability mass we are computing corresponds to the random variable of the accepted price of the customer, which is usually different from the true reservation price of the customer; in the case the customer has a higher reservation price, the hotelier with their offer is "leaving money on the table" as long of course as the next best offer for the auction is higher than the customer's reservation price.

## 5. Algorithms for Forward and Reverse Auction Optimization
### 5.1 Greedy Heuristics for Forward Auction Models

As already mentioned in sections 3.1—3.2, the problem (IP1) is quite non-trivial to formulate mainly due to the flexibility it affords the potential customers in specifying the exact period they wish to come to the hotel. One relatively straight-forward method for this problem is based on the idea of an initial sorting of the various offers made by customers according to some cost function, and the subsequent placement of offers in time as long as there is no overlap with other offers already accepted and schedule in time.

The algorithm works as follows: starting with the best in terms of daily revenue offer, the system decides for which time interval it can commit to accepting the offer. If the currently examined offer cannot be accepted due to luck of a contiguous time interval during which the system can commit to granting to the currently examined bid, the offer is rejected, and the algorithm goes to examine the next best offer. The algorithm stops when it has examined all offers, and for each one it has either been accepted with a specification of the exact time period during which the offer is accepted, or else it has rejected it. The algorithm is depicted in fig. 4 and is based on the function $place(o, R)$ which returns the *first* (in time) interval $\{l, u\} \subseteq \{o.L_{c,r}, o.U_{c,r}\}, |u - l| = o.M_{c,r} - 1$ during which the room-type $r$ is available according to the already accepted and placed offers in the set $R$. If such availability does not exist, the for the interval $\{l, u\}$ we have that $u = l - 1$.

It is easy to see that the above-described algorithm is rather similar to various CPU job scheduling algorithms implemented by Operating Systems following the First-Fit policy, or alternatively, by writing in many sectors of the disk files of certain sizes. All such algorithms suffer from the inevitable fragmentation of the total capacity of a given resource as assignments to the resource increase with time.

> **GREEDY PLACEMENT OF NON-OVERLAPPING OFFERS FOR PROBLEM (IP1)**
>
> **Inputs**: Set of offers $O = \{o = <c, r, D_{c,r} = \{d_{l_{c,r}}, d_{l_{c,r}+1}, \ldots d_{u_{c,r}}\} \subseteq \{L_{c,r}, \ldots U_{c,r}\} \subseteq D, |D_{c,r}| = M_{c,r}, b_{c,r,D_{c,r}}>\}$
>
> **Output**: Set of chosen non-overlapping offers $O'$ with exactly specified time-periods to visit.
>
> 0. **set** $R = \emptyset$.
>
> 1. **sort** the elements $o \in O$ **according** to the quantity $o.b_{c,r,D_{c,r}}$ in **descending** order.
>
> 2. **for-each** $o \in O$ **do**:
>
> 3.   **set** $\{l, u\} = place(o, R)$.
>
> 4.   **if** $u > l$ **then set** $R = R \cup \{<c, r, \{d_l, d_u\}, b_{c,r,D_{c,r}}>\}$.
>
> 5. **end-for**.
>
> 6. **return** $R$.

*Figure 4: Greedy Algorithm for Solving (IP1)*

### 5.2. AI-driven Optimization of Reverse Auction Bid Prices

As an alternative to the expected profit maximization model (MP) of section 4, in the face of historical data that may not match exactly the current reverse auction properties, we use a combination of Data Mining (DM) (Christou et al., 2018) and Machine Learning (ML) (Cohen & Singer (1999), Christou et al., (2022)) techniques to extract all rules that hold with certain support and confidence on the dataset, and then combine them to estimate the price level offer that will maximize the expected profit of the hotelier. The advantage of this approach is due to the fact that it handles missing attribute values by design, and it is a by-default, explainable outcomes approach, meaning that it can (almost) always provide an explanation for its actions, something that most of the competition, including neural networks and deep learning approaches, cannot do without resorting to expensive 3rd party tools and techniques such as Shapley values, LIME etc. In this approach, we model the hotelier's profit maximization problem in a reverse auction as a supervised ML problem, specifically as a regression problem, with the target attribute being a scalar quantity that represents the price target for the offer that is expected to be accepted by the customer that initiated the reverse auction. The independent variables are the various (numeric and/or categorical) attributes that the customer specifies in their reverse auction; example attributes would include request for particular breakfast type (categorical), distance to the sea (numeric), air/conditioning (categorical), check-in period (usually categorical, but could be classified as numeric) etc.

Our AI-driven optimization algorithm is shown in fig. 5.

> **QARMA-based Optimization of Reverse Auction Bid**
>
> **Inputs**: Set $D$ of *accepted* historical offers for reverse auctions, and set of requested characteristics in current reverse auction $A_R$, user-defined support and confidence thresholds $s, c > 0$ and user-defined maximum number $n_a$ of antecedent conditions in the derived rules.
>
> **Output**: Estimated price as response to current reverse auction $A_R$ together with offered characteristics' values.
>
> 1. **run QARMA (Christou et al., 2018)** on the dataset $D$ with parameters $s, c, n_a$ and target variable the price of the accepted historical offer $p$, to produce all rules of the form $I_1 \in [l_1, h_1] \wedge ... \wedge I_m \in [l_m, h_m] \to p \geq v$ and of the form $I_1 \in [l_1, h_1] \wedge ... \wedge I_m \in [l_m, h_m] \to p \leq v$, and put all these rules in the ruleset $R_Q$.
>
> 2. **remove** all rules in $R_Q$ whose antecedents cannot be satisfied by the hotelier's available rooms (e.g. if a rule specifies in the antecedents that the distance from the sea is less than 50m but the hotelier's hotel is located 100m from the sea, remove such a rule)
>
> 3. **run the algorithm (Christou et al., 2022)** on the ruleset $R_Q$ and the set of requested characteristics in the current reverse auction $A_R$ to estimate the optimal price $p^*$ to offer in the current reverse auction.
>
> 4. **choose** among all rules in $R_Q$ for which $p^*$ is an admissible target value, the one whose antecedent conditions, $AC$, incur minimum cost for the hotelier.
>
> 5. **return** the tuple $(p^*, AC)$.

*Figure 5: AI-driven Optimization of Reverse Auction Bid Offers*

## 6. Computational Results
### 6.1 Forward Auction Results

We initially implemented a prototype desktop system that accepts user bids as input data for the forward auction problem, along with auction settings, models the problem in LP format, and solves it using either a state-of-the-art commercial solver (GUROBI 10) or a state-of-the-art Open-Source solver (SCIP 8, or GLPK 5). In fig. 6, we show a prototype Graphical User Interface we implemented that allows I/O between the user and the system. The data required for the formulation of the (IP1') model are entered through the main screen shown in fig. 6, and once all bids are placed, the user hits the "Run" button to start the optimization.

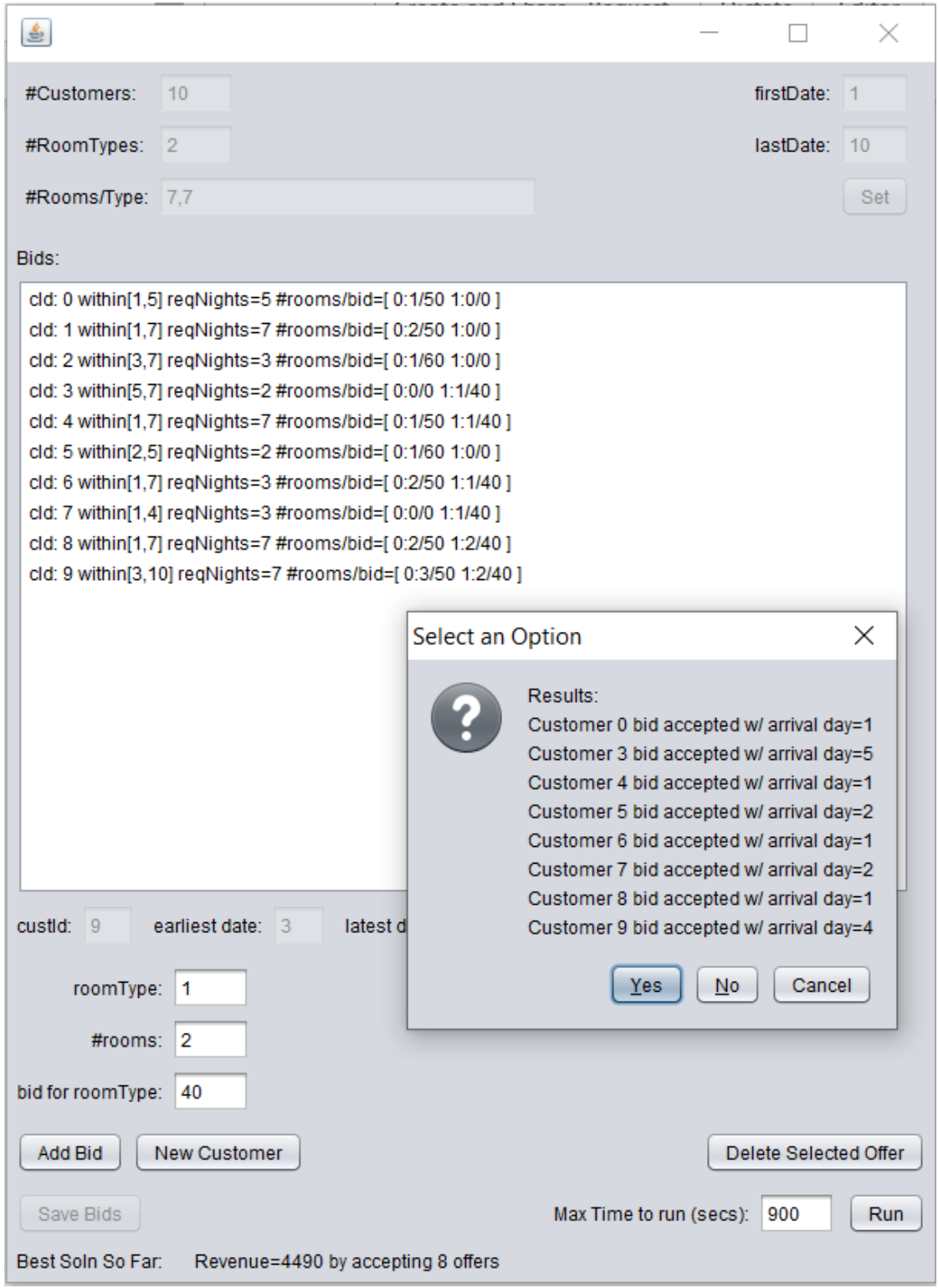

*Figure 6: GUI for Prototype Hotelier Forward Auction System*

The computational results in Table 1 show that (IP1') is easily solvable on a 24-core hyper-threaded PC in near real-time, regardless of the capacity of the hotelier. In Table 1 the first 4 columns show respectively the range of dates for which the auction holds, the capacity of the hotel (in terms of auctioned rooms per room type), the number of customer bids, and the number of bids exceeding capacity for any given day in the range of the auction for any given room type. The last two columns show the response times in seconds of GUROBI and SCIP respectively. Given the near real-time response of the system, there is no need to run the heuristic algorithms of section 5.1, which might only be needed for much larger-scale problems, which however are not of concern for the type of hotel businesses we have in mind (small to medium units.)

*Table 2: Run-times Required for (IP1') for Small to Medium Hotel Units*

| Auction Dates | Capacity per Room Type | #Customers | #Bids Above Capacity per Room Type | Run-time GUROBI (secs) | Run-time SCIP (secs) |
|---|---|---|---|---|---|
| 1-10 | 7,7 | 10 | 6,0 | 0.01 | 0.01 |
| 1-7 | 5 | 10 | 1 | 0.06 | 0.05 |
| 1-14 | 10,5 | 20 | 11,4 | 0.01 | 0.01 |
| 1-30 | 5,5,5 | 20 | 9,9,0 | 0.11 | 0.1 |
| 1-45 | 5,10 | 20 | 0,0 | 0.07 | 0.09 |

As can be seen from the results in Table 2, the run-times of GUROBI and SCIP are very much comparable; this is due to the fact that the techniques used to solve the problem are implemented in both solvers, and are highly efficient, so that it never takes more than a tenth of a second to solve any model created. We estimate that it would take much longer auction period lengths (in the order of 50-60 days), combined with several different room types, each of sufficiently higher capacity (more than 20-30) and many more customers (in the order of 50 or more customers) bidding for the same rooms in order for the solvers to take more than a few seconds to compute the optimal solution to these models. Even though such use cases are not of direct concern in this paper, in order to understand if the optimization technology of today would present a limitation when trying to apply the above ideas to such larger scales, we show in Table 3 below a few results from such larger hypothetical cases.

*Table 3: Run-times Required for (IP1') for Larger Hotel Units*

| Auction Dates | Capacity per Room Type | #Customers | Run-time GUROBI (secs) | Run-time SCIP (secs) |
|---|---|---|---|---|
| 1-60 | 30,30,10,10 | 50 | 0.2 | 1.0 |
| 1-60 | 30,20,20 | 80 | 0.19 | 1.0 |

As can be easily seen however, even in these larger cases (horizon of 60 days with up to 80 potential customers submitting bids) the run-time for solving the corresponding model does not exceed 1 second using the Open-Source SCIP solver; therefore, it is clear that the model (IP1') specified is easily solvable by the current optimization technology on current hardware.

For ease of integration with other services and for maintenance, our initial desktop system design has been supplemented by a RESTful web-service integrated into a web application that fully implement the entire system under consideration. The system responds to HTTP GET requests according to an API we designed as follows.

To solve a forward auction problem that is registered in the database with forward auction id=id we send an HTTP GET Request to the host server with a single parameter being the id of the forward auction:

http://host-server[:port]/api/optimize_auction/{id}

Example:

http://localhost:8080/api/optimize_auction/1

The result of the above call for the particular auction with id 1 is as shown in fig. 7.

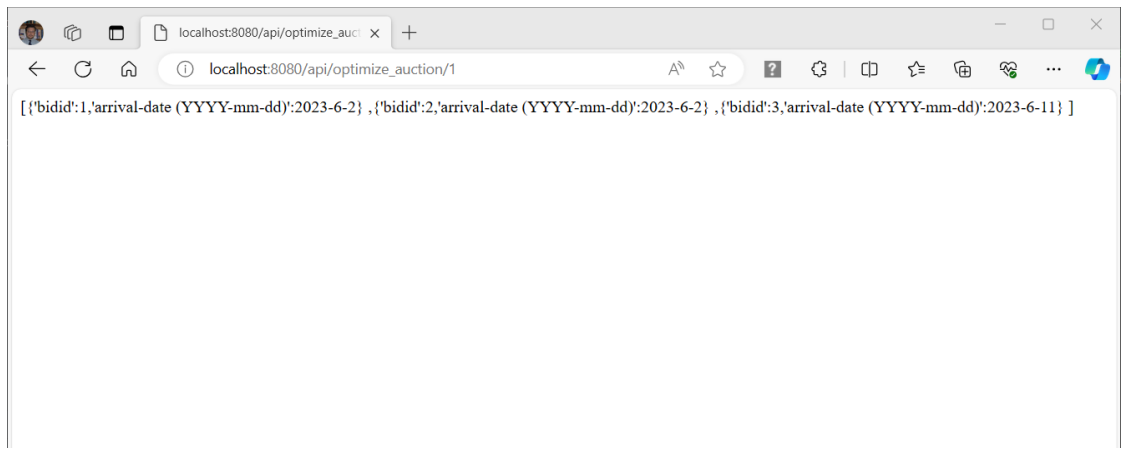

*Figure 7: REST Interface for Forward Auctions*

In this particular example, there are 3 bids that are accepted:

- The bid with database id 1 is accepted and the system specifies the check-in date as June 2, 2023 (2023-6-2)
- The bid with database id 2 is also accepted and the system specifies the check-in date as June 2, 2023 (2023-6-2)
- The final accepted offer is for bid with database id 3, and the required check-in is set for June 11, 2023 (2023-6-11).

Overall, the response to a forward auction problem request is a JSON object as follows:

[

{'bidid':<id>, 'arrival-date (YYYY-mm-dd)':<date>},

…

]

Every dict in the JSON list describes an accepted bid, its id and the check-in date.

## 6.2 Reverse Auction Results

As already discussed in section 4, the hotelier's problem of determining the price to offer for a given reverse auction is simple when historical data of (almost) identical reverse auctions exist and the hotelier has access to them. In such a case, the linear-time scan of the data can be used to determine the price-point that will maximize the expected profit of the hotelier in the long-run, as shown in section 4.1.2.

For the more interesting case where the historical data cover many different cases, many of most of which have only a few variable values in common with the current reverse auction that the hotelier needs to decide upon, we resort to the Data Mining algorithm listed in Fig. 5. As we do not have real-world data for such a case, we resort to a set of synthetic data generated from a set of underlying assumptions that we gathered after an initial set of interviews with the hoteliers' union of an island in the East Aegean. The assumptions are the following: (i) there are 6 main characteristics to monitor:

- "period visiting" (categorical attribute taking values 1=low season, 2=intermediate season, 3=high season),
- "hotel rating" (numeric attribute in [1,5]),
- "distance to sea" (numeric attribute in [0, 10000]),
- "number of beds requested" (numeric attribute in [1,3]),
- "type of breakfast" (categorical attribute taking values 0=N/A, 1=continental, 2=American), and finally
- "number of worth-visiting sites within 10 km" (numeric attribute in [0,10])

(ii) the prices offered for two rooms with the same characteristics (including the ones accepted) cannot differ by more than €30 per night, (iii) all price offers must be in the interval [10, 250] per night, and (iv) prices offered must be monotonic with respect to each numeric variable except for the "distance to sea" variable.

Given the above assumptions, we created a synthetic dataset of 100 reverse auction accepted responses to equal requests, with the goal of computing the total time that the algorithm in fig. 5 takes to compute an optimal reverse auction response to a particular request.

We ran the algorithm on a high-end PC equipped with an intel core i9-10920X processor and 64GB RAM. The user defined support threshold is set at 15% and the corresponding threshold is set at 90%. The maximum number of antecedent conditions in the rules is set at $n_a = 3$. The results of running step 1 (QARMA) of the algorithm are shown in fig. 7.

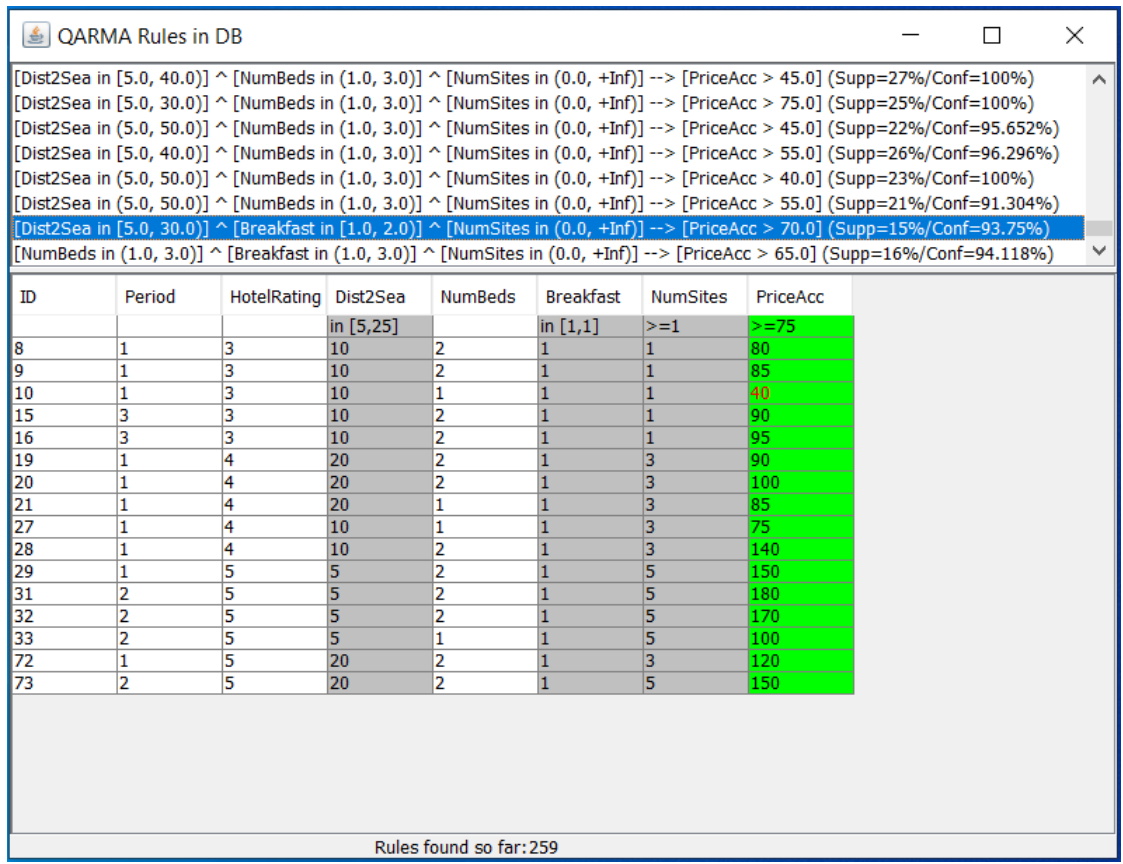

*Figure 8: Rules Produced by Running QARMA on the Synthetic Dataset for Reverse Auction Problems: the selected rule states that when the requested distance-to-the-sea is between 5 and 25 meters, the requested breakfast type is continental, and the number of near-by sites is at least 1, the accepted price is at least €75 per night. The bottom part of the GUI shows the list of all instances in the database that satisfy the antecedents of the rule.*

Results from running the entire algorithm in fig. 5 are shown in Table 4 for 10 different unseen before (synthetic) reverse auction requests. In the "Est. Price Response" column we show the results of the algorithm estimation, together with the "ground truth" acceptance price. When the acceptance price was lower than the estimated price the result is shown in red, as it indicates that the estimate price would lead to a losing offer (the client wouldn't accept it.) As can be seen, running the entire algorithm takes no more than a few *milli-seconds* of wall-clock time, making the entire system interactive in near real-time.

*Table 4: Data Mining Algorithm Results for Estimating Optimal Price in Reverse Auctions*

| Period Visiting Req. | Hotel Rating Req. | Distance to Sea Req. | #Beds Req. | Breakfast Type Req. | #sites Req. | Est. Price Response | Response Time (msecs) |
|---|---|---|---|---|---|---|---|
| 1 | 4 | 30 | 1 | 1 | 0 | 60(40) | 57 |
| 1 | 4 | 30 | 1 | 2 | 1 | 80(50) | 50 |
| 3 | 3 | 25 | 2 | 2 | 2 | 80(100) | 55 |
| 3 | 2 | 50 | 3 | 1 | 3 | 80(90) | 57 |
| 1 | 2 | 50 | 2 | 2 | 1 | 70(50) | 55 |
| 1 | 1 | 200 | 1 | 1 | 0 | 37.7(30) | 54 |
| 1 | 1 | 200 | 2 | 1 | 0 | 40(50) | 59 |
| 1 | 1 | 200 | 2 | 2 | 0 | 56(60) | 50 |
| 1 | 2 | 100 | 1 | 1 | 1 | 50(55) | 51 |
| 1 | 3 | 800 | 2 | 3 | 4 | 60(70) | 50 |

The Mean Absolute Error of the system predictions is quite low, at €13.11, which means that the estimated acceptance price is on average only 13.11 Euros away from the true acceptance price! On the other hand, the Mean Absolute Percentage Error of the system is 25.6%; this percentage error would definitely be improved a lot by a larger database containing a few hundred to a few thousand data points; more attributes (such as finer-grain periods) including extra characteristics of the hotel such as tennis courts, heated pools etc, as well as amenities in the rooms, such as existence of jacuzzi baths, private pools etc. would also allow for much more accurate estimation of the acceptance price of a user to a reverse auction.

## 7. Conclusions and Future Directions

We have introduced forward and reverse auctions to balance offer and demand for hotel reservations ahead of time. In the forward auction case, the hotelier auctions one or more of their available rooms during a period of their choice, and customers may place sealed bids over the internet for one or more of the hotelier's rooms. The customers are also allowed to specify a minimum and maximum date during which they would like to visit, together with the exact number of nights they would like to spend in the hotel. Once the period during which auctions may be placed passes, the hotelier needs to choose among the available bids (offers), the ones that together maximize their profits. We model the problem as a deterministic Mixed Integer Programming model, and show that for all cases of interest, the problem can be solved using State-of-the-Art Open-Source solvers in almost real-time.

The reverse auction problem is the problem where a potential visitor of a place posts a reverse auction request, stating their intention to visit the place in certain period, and asks near-by hoteliers to make an offer for staying in their hotel; the customer sets their requirements, and the hotelier has to decide whether to make an offer or not; if the hotelier decides to make the customer an offer, they need to decide also the price for their offer. We model this problem as a stochastic optimization problem where the hotelier needs to specify the price level of their offer so as to maximize their expected profit. As long as the expected profit is positive, the hotelier proceeds with the offer, whereas in case of non-positive expected profit, the hotelier should decide to abstain from the reverse auction process. We have shown that in cases where historical data matching the requirements of the reverse auction exist, the stochastic optimization problem is directly solvable in linear time (the time required to scan the historical data once); in cases where no historical data match the current reverse auction requirements exactly, a data mining algorithm that extracts all quantitative association rules that hold on a dataset is employed to pick the cases that most closely match the current one.

We are already in the process of integrating the prototype optimization and data mining systems we have into a full cloud-based web application that will allow hoteliers of lesser known tourist areas and visitors of these areas alike to benefit from the above match-making systems so as to get a more complete and satisfying tourism experience.

## Acknowledgements
This research has been co-financed by the European Regional Development Fund of the European Union and Greek national funds through the Operational Program Competitiveness, Entrepreneurship and Innovation, under the call RESEARCH-CREATE-INNOVATE (project code: T2EΔK-00726)